%% file: main.tex
\documentclass{article}

\usepackage{microtype}
\usepackage{graphicx}
\usepackage{subfigure}
\usepackage{booktabs} 

\usepackage{multirow}
\usepackage{makecell}
\usepackage{caption}
\usepackage{subcaption}
\usepackage{siunitx}

\usepackage{hyperref}

\usepackage[accepted]{icml2024}

\usepackage{amsmath}
\usepackage{amssymb}
\usepackage{mathtools}
\usepackage{amsthm}

\usepackage[capitalize,noabbrev]{cleveref}

\theoremstyle{plain}

\theoremstyle{definition}

\theoremstyle{remark}

\input{preamble}

\usepackage[textsize=tiny]{todonotes}

\icmltitlerunning{Dense Dynamics-Aware Reward Synthesis}

\begin{document}

\twocolumn[
\icmltitle{Dense Dynamics-Aware Reward Synthesis:\\Integrating Prior Experience with Demonstrations}



\icmlsetsymbol{equal}{*}

\begin{icmlauthorlist}
\icmlauthor{Cevahir Koprulu}{equal,ut}
\icmlauthor{Po-han Li}{equal,ut}
\icmlauthor{Tianyu Qiu}{ut}
\icmlauthor{Ruihan Zhao}{ut}
\icmlauthor{David Fridovich-Keil}{ut}
\icmlauthor{Sandeep Chinchali}{ut}
\icmlauthor{Ufuk Topcu}{ut}
\icmlauthor{Tyler Westenbroek}{uw}
\end{icmlauthorlist}

\icmlaffiliation{ut}{University of Texas at Austin, USA}
\icmlaffiliation{uw}{University of Washington, USA}

\icmlcorrespondingauthor{Cevahir Koprulu}{cevahir.koprulu@utexas.edu}
\icmlcorrespondingauthor{Po-han Li}{pohanli@utexas.edu}

\icmlkeywords{Reinforcement Learning, Learn from Demonstrations, Online Reinforcement Learning, Imitation Learning}

\vskip 0.3in
]



\printAffiliationsAndNotice{\icmlEqualContribution. Under Review. Code: \url{https://github.com/cevahir-koprulu/dense_demos/tree/main}} 

\begin{abstract}
Many continuous control problems can be formulated as sparse-reward reinforcement learning tasks. In principle, online reinforcement learning methods can automatically explore the state space to solve each new task. However, discovering sequences of actions which lead to a non-zero reward becomes exponentially more difficult as the task horizon increases. Manually shaping rewards can accelerate learning for a fixed task, but it can be an arduous process that must be repeated for each new environment. This work introduces a systematic reward-shaping framework which distills the information contained in $1)$ a task-agnostic prior data set and $2)$ a small number of task-specific expert demonstrations, and then uses these priors to synthesize dense dynamics-aware rewards for the given task. This supervision substantially accelerates learning in our experiments, and we provide analysis demonstrating how the approach can effectively guide online learning agents to faraway goals. 
\end{abstract}

\section{Introduction}
\label{sec:intro}

\input{intro}

\section{Problem Setting}
\label{sec:prelim}
\input{prelim}

\section{Systematic Reward-Shaping from Demonstrations and Prior Experience}
\label{sec:method}
\input{method}

\section{Experiments}
\label{sec:expirements}
\input{expirements}

\section{Discussion}
\label{sec:discussion}
\input{discussion}

\clearpage
\section*{Impact Statements}
This paper presents work whose goal is to advance the field of Machine Learning. There are many potential societal consequences of our work, none which we feel must be specifically highlighted here.

\bibliography{refs}
\bibliographystyle{icml2024}

\clearpage
\onecolumn
\appendix
\section{Appendix}
\label{sec:appendix}
\input{appendix}

\end{document}

%% file: preamble.tex
\usepackage{xcolor}
\usepackage{etoolbox}
\usepackage{upgreek}
\usepackage{amsmath}
\usepackage{comment}

\usepackage{amssymb}
\usepackage{mathtools}
\let\set\relax
\usepackage{booktabs}
\usepackage{multirow}
\usepackage{siunitx}

\usepackage{caption}
\usepackage{subcaption}
\usepackage{graphicx}
\usepackage{bm}
\usepackage{bbm}
\usepackage{enumerate}
\usepackage{wrapfig}

\definecolor{forestgreen}{RGB}{34,139,34}

\newtheorem{prop}{Proposition}

\newcommand{\set}[1]{\{ #1\}}

\DeclareMathOperator*{\argmax}{arg\,max}

\newcommand{\R}{\mathbb{R}}

\newcommand{\plnote}[1]%
    {\textcolor{orange}{\textbf{PHL: #1}}}
\newcommand{\twnote}[1]%
    {\textcolor{cyan}{\textbf{TW: #1}}}
\newcommand{\aanote}[1]%
    {\textcolor{blue}{\textbf{#1}}}
\newcommand{\ksnote}[1]%
    {\textcolor{red}{\textbf{KS: #1}}}    
\newcommand{\qtynote}[1]%
    {\textcolor{green}{\textbf{QTY: #1}}} 
\newcommand{\david}[1]%
    {\textcolor{blue}{\textbf{David: #1}}}
\newcommand{\rz}[1]%
{\textcolor{red}{\textbf{Philip: #1}}}

\newlength\tindent
\newcommand{\msnote}[1]%
    {\textcolor{cyan}{\textbf{MS: #1}}}

%% file: intro.tex
\input{figure_latex/system_diagram}

Intelligent decision-making requires temporally extended reasoning across a multitude of environments. In principle, established exploration techniques for online deep reinforcement learning (RL) \citep{haarnoja2018soft,schulman2017proximal, chentanez2004intrinsically, bellemare2016unifying} can achieve the spatial data coverage needed to carry out each new task. However, exploration for long-horizon tasks, especially those with sparse rewards, remains a significant bottleneck. In contrast, offline RL attempts to bootstrap learning from diverse data sets \citep{kumar2020conservative, janner2021offline}. While both paradigms can benefit from amortizing learning across many tasks \citep{chane2021goal,andrychowicz2017hindsight}, solving novel long-horizon tasks remains challenging. 

Learning from Demonstrations (LfD) frameworks can leverage expert trajectories to learn effective actions over long horizons. However, they struggle to obtain the necessary spatial data coverage to learn robust policies. This is due to compounding errors, wherein small differences in initial conditions or actions cause rollouts to rapidly diverge from the expert training distribution \citep{ke2023ccil,ross2011reduction, block2023provable}. Thus, LfD techniques may require an impractical number of expensive expert demonstrations \citep{ross2011reduction}. Low-quality or sub-optimal demonstrations exacerbate this challenge, degrading the performance of existing methods.

This paper introduces a novel reward-shaping framework that embraces the strengths of these paradigms. Our approach, see \cref{fig:system_diagram}, combines task-agnostic prior experience with a handful of task-specific expert demonstrations to synthesize dense dynamics-aware rewards that substantially accelerate online exploration and learning. Our approach encodes information about the dynamics during an initial pre-training phase by learning a goal-conditioned value function from task-agnostic data. Then, given a few temporally extended demonstrations for a new task, we leverage the potential-based reward-shaping (PBRS) formalism \citep{ng1999policy} to construct a potential with two terms: $1)$ a term incentivizing the agent to reach the nearest point on a demonstration, via the goal-conditioned value function, $2)$ a term encouraging the agent to follow the demonstration from this point to the task-specific goal. The shaped reward provides dense supervision, biasing the exploration towards nearby expert demonstrations to faraway goals. 
In summary, our approach leverages expert demonstrations for temporally extended reasoning, task-agnostic data to `fill the gaps' around demonstrations with dense rewards, and online learning to smooth over potential deficiencies in these data sources. 
Beyond accelerating learning, our approach has the following benefits: 

\textbf{Overcoming Sub-optimal Demonstrations and Dynamics Shift:} The potential in PBRS can be viewed as an approximation to the optimal value function of the original objective \citep{ng1999policy}. While the reshaped reward guides the exploration, the new objective preserves the optimal policies for the original problem. In our setting, the potential will be a sub-optimal approximation when the demonstrations are poor. Similarly, when dynamics shift between the prior data and the target task, the goal-conditioned value functions may be inaccurate. Nonetheless, as our experiments demonstrate, our approach still accelerates learning optimal policies in these scenarios. 

\textbf{Learning from (Partial) Observations:} In practice, the agent may not have access to the expert's actions, only partial observations, and may even need to infer states from high-dimensional observations such as visual inputs. For example, when demonstrating how to push an object across a table, a human expert may use their hands rather than teleoperating a robot arm. Our reward-shaping approach only requires the expert's states, and we demonstrate how the approach extends to partial (and potentially high-dimensional) observations. However, because we implement our approach with off-policy RL, when actions are available, we benefit from including expert transitions in the replay buffer \citep{nair2018overcoming, rajeswaran2017learning}.
 
\textbf{Shortening Learning Horizons:} During the initial pre-training phase, we focus on learning accurate goal-conditioned value estimates for goals that take fewer steps to reach in comparison to temporally extended expert demonstrations. Shortening the learning horizon enables us to leverage tractable amounts of prior data effectively while reusing it for many downstream tasks. We also investigate the effects of annealing the discount factor \citep{cheng2021heuristic} for the reshaped objective. Theoretically and empirically, we show how using a small discount factor early in training will cause the agent to greedily exploit the prior information baked into the dense rewards during initial exploration. Our experiments indicate this can further accelerate learning by a substantial margin. 


\section{Related Work}

\textbf{Overcoming Distribution Shift in Imitation Learning:} A challenge for imitation learning is distribution shift, wherein small errors compound over time and cause the state distribution generated by the agent to diverge from that of the expert. Techniques such as DAGGER \citep{ross2011reduction} and DART \citep{laskey2017dart} use an interactive expert model to query expert trajectories that recover from mistakes made by the agent; however, this process may be overly arduous or impossible to implement in many practical scenarios. Frameworks such as GAIL \citep{ho2016generative}, SQIL \citep{reddy2019sqil}, and AIRL \citep{fu2017learning} allow the agent to interact with the environment independently and attempt to match the expert distribution over long horizons. The approach in \citep{reichlin2022back} learns a recovery policy that takes the agent back to the set of demonstrated states but requires that the dynamics are known in the testing environment. Generative techniques such as \citep{ke2023ccil} use a learned model to perform data augmentation by generating actions returning the agent to the expert distribution. While these approaches are highly effective at imitating the expert, their performance is fundamentally limited by the quality of the demonstrations. 

\noindent\textbf{Reinforcement Learning with Demonstrations:} Most similar to our approach are techniques that use expert demonstrations to accelerate base off-policy RL methods. Approaches such as \citep{nair2018overcoming, rajeswaran2017learning, hester2018deep} add demonstrations to replay buffers of off-policy RL methods and use techniques such as prioritized experience replay and pre-training with behavior cloning to accelerate and stabilize learning. These methods are directly compatible with our core reward-shaping technique. Other contributions such as \citep{brys2015reinforcement, zhao2023learning, pari2021surprising} have performed reward-shaping to guide agents back to demonstrated states but rely on heuristic hand-designed measures of the distance between a given state and a nearby demonstration (such as simple L2 distances). \citet{wilcox2022monte} use suboptimal but task-specific demonstrations. Our method bakes task-agnostic prior knowledge about the dynamics into the reward, which, as we discuss in Sec. \ref{sec:discounting}, is essential to accelerate learning.

%% file: figure_latex/system_diagram.tex
\begin{figure*}
    \centering
    \includegraphics[width=0.85\textwidth]{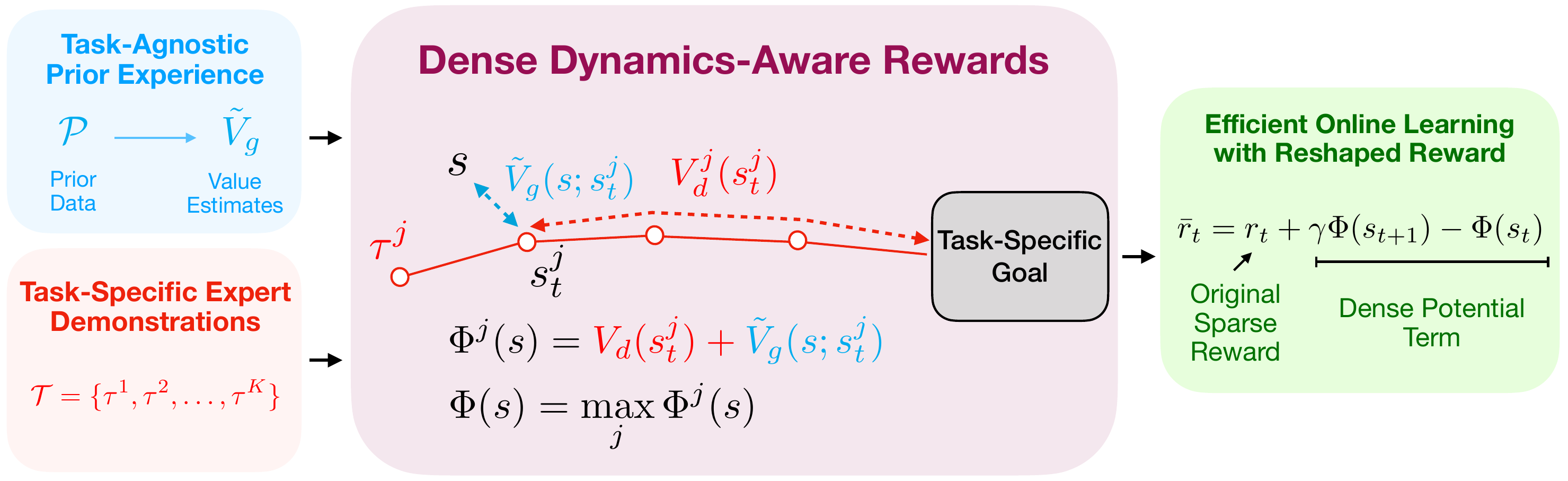}
    \caption{\small{A depiction of our framework. ({\color{cyan}Cyan}) We distill information about the dynamics of a prior data set into a goal-conditioned value estimator $\tilde{V}_g$. ({\color{red}Red}) An expert provides demonstrations for a new target task. ({\color{purple}Purple}) The framework combines them to construct a potential function $\Phi(s)$ \eqref{eq:sinle_pot}, implicitly estimating the number of steps needed to reach the task-specific goal from state $s$. For this estimate, $\Phi^j(s)$ first measures \eqref{eq:demo_pot} the number of steps needed to reach the $j$-th demonstration $\tau^j$ via  $\tilde{V}_g$, and then follow it to the goal via $V_d^j$ \eqref{eq:demo_reward}. ({\color{forestgreen}Green}) We use the overall potential to synthesize dense dynamics-aware rewards for the target task.}}
    \label{fig:system_diagram}
\vspace{-2em}
\end{figure*}

%% file: prelim.tex
\subsection{Sparse Reward Reinforcement Learning Problem}
Our ultimate objective is to efficiently solve long-horizon sparse reward reinforcement learning problems defined by an MDP $\mathcal{M} = (\mathcal{S},\mathcal{A}, P, r, \gamma)$ where $s \in \mathcal{S}$ is the state, $ a \in \mathcal{A} \subset \R^a$ is the action, $s_{t+1} \sim P(\cdot |s_t,a_t)$ is the transition dynamics, $r(s,a)$ is the rewards, and $\gamma \in [0,1)$ is the discount factor. We will focus on the case where the agent aims to reach a goal set of states $\mathcal{G} \subset \mathcal{S}$ in as few steps as possible by optimizing over $\Pi$, the space of all policies $\pi$.
For ease of analysis, we assume that the dynamics and policies are deterministic to simplify the exposition, as in prior works studying the structure of goal-conditioned problems \citep{ma2022vip}. Namely, $P$ can be described by a single-valued map $f$ such that $s_{t+1} =f(s_t,a_t)$ and actions are generated according to $a_t =\pi(s_t)$. 
However, our method also applies to stochastic scenarios, as demonstrated in our experiments with stochastic policies.
To incentivize the desired goal-reaching behavior, we formulate a policy optimization problem of the form:
\begin{equation}
    \max_{\pi \in \Pi}~ \mathbb{E}_{s_0 \sim d_0}V^\pi(s_0), ~~~
    V^\pi(s_0) := \sum_{t=0}^\infty \gamma^tr(s_t,a_t),
\end{equation}
where $d_0$ is a distribution over initial conditions. We use a sparse reward of $r(s_t,a_t) = 1$ if $s_{t+1} \in G$ to award successes and apply a reward of $r(s_t, a_t)=0$ otherwise. Episodes terminate when the agent first reaches $\mathcal{G}$. The larger the magnitude of $V^\pi(s)$, the fewer steps it will take $\pi$ to reach the goal set $\mathcal{G}$ when starting from $s$ and using $\pi$. We define the optimal value function as:
\begin{equation}
V^*(s_0)  := \max_{\pi \in \Pi} V^{\pi}(s_0).
\end{equation}
The magnitude of $ V^*(s)$ captures the optimal number of steps needed to reach the goal from $s$.

\subsection{Goal-conditioned Policies}
Consider a goal-conditioned controller of the form $a_t = \pi_g(s; s_d)$ where $s \in \mathcal{S}$ is the current state, and $s_d$ is the desired state the agent attempts to reach. Let $\Pi_g$ denote the space of all deterministic goal-conditioned policies. For each policy $\pi_g$, we define the value function:
\begin{equation}
V_g^{\pi_g}(s_0;s_d) := \sum_{t =0}^{\infty} \gamma^{t} r_g(s_t, a_t;s_d),
\end{equation}
where we use the sparse rewards $r(s_t, a_t;s_d) =1$ if $s_{t+1} = s_d$ to award the agent for successfully reaching the desired state, and apply a reward of $r(s_t,a_t;s_d) = 0$ otherwise. Episodes for the goal-conditioned agent terminate once the agent reaches the goal state and the magnitude of $V_g^{\pi_g}(s;s_d)$ captures the number of steps the given controller will take to reach the goal $s_d$ from $s$.

%% file: method.tex

\subsection{Learning Approximate Goal-conditioned Value Estimates from Prior Data}
Our approach begins with a pre-training phase, wherein the agent leverages a broad prior data set of transitions $\mathcal{P} = \set{(s_t^p,a_t^p,s_{t+1}^p)}_{t=1}^{T_p}$, which can be generated by one or multiple prior MDPs where data have been collected. We assume that each of these MDPs has the same state and action space as the target MDP $\mathcal{M}$, but the dynamics of the prior MDPs can differ substantially from those encountered in $\mathcal{M}$. For example, in our experiments in Sec. \ref{sec:expirements}, the prior data set comes from an environment without a slot.
We use $\mathcal{P}$ to fit an approximation $\tilde{V}_g(s;s_d)$ to the goal-conditioned value function $V_g^{\pi_g}$ associated to some fixed policy $\pi_g$ (but not necessarily the optimal goal-reaching policy). Our framework is agnostic to how $\tilde{V}_g$ is learned. For example, offline RL can be used to learn both $\pi_g$ and $\tilde{V}_g$ if $\mathcal{P}$ is a static data set \citep{mezghani2023learning}. Or, as in our experiments, these quantities can be learned by generating $\mathcal{P}$ using online exploration and extensive play data. 

Regardless of how $\tilde{V}_g$ is learned during the pre-training phase, it encodes valuable information about the dynamics that generated the prior data that we will use to accelerate learning for the downstream task defined by the target MDP $\mathcal{M}$. However, $\tilde{V}_g$ may be flawed due to $1)$ dynamics shift between the prior training data and the target environment and $2)$ limited data coverage in $\mathcal{P}$. In particular, $2)$ can make obtaining accurate value estimates for faraway goals extremely difficult. Thus, in practice, we aim only to learn value estimates for goals that require a relatively small number of steps to reach, reducing the horizon and sample complexity for the initial pre-training phase. For example, in our pushing experiments, we train controllers that can push a puck to goals up to $0.1$ meter away during the pre-training phase and then use these primitives to track much longer expert demonstrations to distant goals. 
For concreteness, for each starting state $s \in \mathcal{S}$, let $\Delta(s) \subset \mathcal{S}$ denote the range of goals that we use to fit $\tilde{V}_g$ during the pre-training phase. For our analysis, it is convenient to assume that the sets of valid goals are of the form:
\begin{equation} \label{eq:goals}
\Delta(s) := \{s_d \in S : \tilde{V}_g(s;s_d) \geq \beta \},
\end{equation}
where $\beta \in [0,1)$ is uniform across all states.

We also highlight related works such as \citep{eysenbach2019search}, which explicitly attempt to estimate whether certain goals are out of distribution for goal-conditioned value estimators. While defining $\Delta(s)$ for the examples we consider is straightforward, these approaches are compatible with our method when it is more difficult to ascertain whether certain states are out of distribution. 

\subsection{Collecting Task-Specific Demonstrations}
Our method assumes access to $K$ expert demonstrations $\tau_d^j$ for the target MDP $\mathcal{M}$. When introducing our base method, we will initially assume that the learner has full access to the transitions performed by the expert. Thus, the expert data set will be of the form: $\mathcal{T} = \{\tau_d^1, \tau_d^2, \dots, \tau_d^K \}, \tau_d^j = \set{(s_t,a_t,s_{t+1}^j)}_{t=0}^{H^j}$,
where $H^j$ is the number of transitions in the $j$-th demonstration. We assume that $s_{H_j + 1}^j \in \mathcal{G}$ and $s_{t}^j \not \in \mathcal{G}$ for each $t < H_j+1$, so that the expert first reaches the goal set.

\subsection{Sythesizing Dense Rewards}\label{sec:reward}
We now introduce our reward-shaping method, which uses the prior experience encapsulated in $\tilde{V}_g$ to synthesize dense dynamics-aware rewards for the target MDP $\mathcal{M}$. We begin by defining:
\begin{equation}\label{eq:demo_reward}
V_{d}^j(s_t^j) := \gamma^{H_j - t},
\end{equation}
This is simply the discounted sum of rewards incurred in the original MDP $\mathcal{M}$ by starting at state $s_t^j$ and following the $j$-th expert demonstration to $G$. This value captures the number of steps needed to reach the goal by following the demonstration. Next, we define:\footnote{We apply the convention that $\Phi^j(s) = 1$ if $s \in G$ and $\Phi^j(s) = 0$ if there does not exist a state $s_t^j$ on the demonstration such that $s_t^j \in \Delta(s)$ (\textit{i.e.}, if the $j$-th demonstration is too far away from $s$).}
\begin{equation}\label{eq:demo_pot}
\Phi^j(s) := \max_{s_t^j \in \Delta(s)}\left[V_d^j(s_t^j) + \tilde{V}_g(s;s_t^j) \right],
\end{equation}
where the maximization is taken over the set of states that appear in $\tau^j$ and are in-distribution goals for $\tilde{V}_g$ for the starting state $s$. It balances choosing states on the demonstration, which will take a small number of steps to reach from $s$ (as measured by $\tilde{V}_g(s;s_t^j)$) against picking states that are closer to the end of the demonstration (as measured by $V_d^j(s_t^j)$). We can view $\Phi^j$ as an estimate of the optimal value $V^*$, which is constructed by approximating how many steps it will take to reach $\tau^j$ and then follow the demonstration to $G$. We aggregate these estimates by defining:
\begin{equation}\label{eq:sinle_pot}
\Phi(s) := \max_{j} \Phi^j(s),
\end{equation}
which captures the optimal demonstration to reach and then follow from the state $s$. We then leverage the PBRS formalism \citep{ng1999policy} and define the dense rewards:
\begin{equation}\label{eq:reshaped_reward}
\bar{r}(s_t, a_t) = r(s_t,a_t) + \gamma \Phi(s_{t+1}) - \Phi(s_t), 
\end{equation}
and propose the reshaped policy optimization problem:
\begin{equation} 
\max_{\pi \in \Pi} \mathbb{E}_{s_0 \sim d_0}\left[\bar{V}^\pi(s)\right], ~~~
\text{s.t. }
\bar{V}^\pi(s) = \sum_{t=0}^{\infty} \gamma^{t}\bar{r}(s_t,a_t),
\label{eq:reshaped_value}
\end{equation}
along the dynamics of the target environment. $\Phi$ is referred to as a \emph{potential function} and can be viewed as an initial guess for the optimal value $V^*$ injected into the reshaped reward structure to guide exploration. In particular, the new rewards incentivize choosing actions that increase the value of $\Phi$ at the following state, guiding learning agents to reach and follow the expert training distribution. This supervision provides a much stronger learning signal than the original spare rewards, which only provides feedback to the agent when it successfully reaches the goal set $\mathcal{G}$. It is extremely difficult, especially when the goal takes many steps to reach.

However, the PBRS formulation preserves the set of optimal policies between the original and reshaped MDPs \citep{ng1999policy}. Thus, this formulation enables us to leverage prior experience and a small number of demonstrations to guide training without tying the end performance of the policies we learn to these potentially flawed data sources. Indeed, as our experiments demonstrate, our approach can still substantially accelerate learning even when highly sub-optimal demonstrations are used and there is a significant dynamics shift between the pre-training data set and target domain.  

\subsection{Guiding Exploration by Decreasing and Annealing Discount Factors}\label{sec:discounting}
To see why the reshaped objective preserves the optimal policies defined by the original MDP, recall that for a given policy $\pi \in \Pi$, the associated $Q$-functions in the original and reshaped MDPs are: 
\begin{equation}
Q^\pi(s,a) =\gamma V^\pi(s') + r(s,a), ~~~
\bar{Q}^\pi(s,a) =\gamma \bar{V}^\pi(s') + r(s,a) + \gamma \Phi(s') - \Phi(s).\label{eq:reshaped_q}
\end{equation}
By unrolling the expression for the reshaped value function \eqref{eq:reshaped_value}, one may show that $\bar{V}^{\pi}(s) = V^\pi(s) - \Phi(s)$. 
Combining \eqref{eq:reshaped_value} and \eqref{eq:reshaped_q}, $\bar{Q}^\pi(s,a) = \gamma V^\pi(s') + r(s,a) -\Phi(s) = Q^\pi(s,a) -\Phi(s)$. Because the $-\Phi(s)$ term does not depend on the action taken by the agent, $\bar{Q}^\pi$ preserves the ordering over actions defined by $Q^{\pi}$. When we take the given policy to be the optimal policy $\pi^*$, this argument shows that the reshaped MDP preserves the optimal ordering of actions in the original MDP. However, in practice, many reinforcement learning algorithms use an approximation to the $Q$-function learned from experience. When these data are limited, and the estimate for the $Q$-function is inaccurate, the shaped rewards can bias the agent towards actions that follow the potential, which enables the reshaped rewards to guide exploration early in the learning process.

We can more aggressively and directly incentivize the agent to follow the potential during exploration by altering the discount factor used during online training \citep{westenbroek2022lyapunov, cheng2021heuristic}. In particular, suppose that we instead apply the long-term return
$\bar{V}^\pi(s_0) = \sum_{t=0}^{\infty}\bar{\gamma}^t \bar{r}(s,a),$
where $\bar{\gamma} \in [0,\gamma]$ is a smaller discount factor (but we leave the original discount factor $\gamma$ unchanged when it is used to construct $\Phi$ and $\bar{r}$). In general, reducing the discount factor for an MDP shortens the effective time horizon over which the agent searches while learning and can reduce the sample complexity and variance of many search methods \citep{westenbroek2022lyapunov, cheng2021heuristic}. However, this regularization may also cause the agent to behave too myopically to achieve the desired long-horizon task. In our setting, the addition of the potential function in the reshaped rewards can guide the agent to act effectively over long horizons even when acting greedily. Consider the $Q$-function for the reshaped objective under the new discount factor
$\bar{Q}^\pi(s,a) = \bar{\gamma}\bar{V}^\pi(s') + \gamma \Phi(s') - \Phi(s) + r(s,a).$
Because the original rewards are sparse, at most states, the only terms affected by the agent's actions are $\bar{\gamma}\bar{V}^\pi(s')$ and $\gamma \Phi(s')$. As $\bar{\gamma}$ gets smaller, the agent prefers to follow the potential over the value function (which must be bootstrapped from data in practice). In the extreme case where $\bar{\gamma} = 0$, the agent greedily follows the potential, relying on the prior information baked into $\Phi$ to guide exploration.
It may be advantageous early in training when data coverage in the new domain is limited (and bootstrapped estimates for $\bar{V}^\pi(s')$ are poor) but will tie the final performance of the policy to the quality of the data used to synthesize $\Phi$. 

One compromise we leverage in our experiments is to train the agent with an increasing sequence of discount factors $\set{\bar{\gamma}_k}_{k=1}^{M}$ where $\bar{\gamma}_1 =0$ and $\bar{\gamma}_M = \gamma$.
By annealing the discount factor, the agent relies more heavily on the potential for guidance when encountering a new environment but still recovers the optimal policy for the environment at convergence. The following result provides sufficient conditions to ensure the online learner can discover trajectories that reach the goal by greedily following the potential. The supplementary material provides the proof in the Appendix \ref{sec:app_proof}. 

\begin{prop}\label{prop}
Assume each state $s \in \mathcal{S}$, in-distribution goal $s_d \in \Delta(s)$, and value estimator $\tilde{V}_g$ satisfies the following conditions for some $\epsilon>0$: \\
\quad 1. $\tilde{V}_g(s;s_d) \in [0,1]$ and  $\tilde{V}_g(s;s_d) =1$ iff $s =s_d$; and,\\
\quad 2. $\max_a\tilde{V}_g(f(s,a);s_d) > (1+\epsilon)\tilde{V}_g(s;s_d)$.\\
Further assume that the expert training set $\mathcal{T}$ satisfies the following condition: \\
\quad 3. for each state $s \in \mathcal{S}$ there exists some state $s_t^j$ on some demonstration $\tau^j$ such that $s_t^j \in \Delta(s)$.
Finally, assume that $\bar{\gamma} = 0$ is used to optimize the reshaped rewards, and let $\bar{\pi}^*$ denote the optimal policy. Then, every trajectory generated by $\bar{\pi}^*$ will reach the goal set $G$ in a finite number of steps. 
\end{prop}
The first assumption in the proposition ensures that the value estimator $\tilde{V}_g$ has a well-defined maximum at the goal state. The second condition says that we can reach each in-distribution goal for the state $s$ by greedily following $\tilde{V}_g$ at each time step. Note that this does not imply that $\tilde{V}_g$ accurately predicts the number of steps needed to reach $s_d$, but rather that it captures the rough structure of the dynamics in the target MDP. Finally, the third condition requires that at every state $s$, there exists an in-distribution goal for $\tilde{V}_g$, which lies on an expert demonstration. In particular, this condition captures the fact that there is a ``tube" around each expert demonstration on which our dense rewards provide useful information (with the radius of the tube corresponding to the range of in-distribution goals for $\tilde{V}_g$). The third condition implies that these tubes cover the entire state space, ensuring sufficient data coverage from the demonstrations and the prior data set to rely on greedily following the potential at every point in the state space. However, while these conditions are strong, we emphasize that we can always ensure that our approach recovers the optimal policy at convergence by annealing the discount factor back to the original value during training. 

\subsection{Learning From (Partial) Observations}
\label{sec:IS}
While our general reward-shaping technique can be incorporated with many paradigms, in this paper, we apply this method in the context of online off-policy RL. Like many other approaches, if the learner can directly observe the actions and states that occur in the expert demonstrations, then the demonstrations can be added to the replay buffer of the off-policy RL algorithm \citep{nair2018overcoming, rajeswaran2017learning}. Importance sampling can replay the demonstrations more frequently, which can help propagate reward information over long horizons.

However, note that our reward-shaping approach, as introduced in Sec. \ref{sec:reward}, only requires access to the states generated by the expert and thus can be applied when only state trajectories are available. Indeed, in our experimental results in Sec. \ref{sec:expirements}, we observe that applying our reward shaping approach without access to expert actions can lead to faster learning when compared to an agent that has access to the expert actions but does not benefit from our informative reward structure.

Our approach extends to scenarios where the expert only has access to partial state information or must infer certain states from high-dimensional observations. Rather than specifying goals with full desired states, in many scenarios, only a subset of states is crucial for defining goals. For example, in many manipulation tasks, success only depends on the final state of the object but not the manipulator. Moreover, in many practical scenarios, such as when a robot observes a human manipulating an object, only the state of the object is observable. Still, there are no concrete actions or states of the robot that can be directly associated with the demonstration.

We can formalize this scenario by breaking down the state into $s =(s_1,s_2)$, where $s_1$ are the goals-relevant states, and we define goal sets of the form $g(s_d^1) = \set{s = (s_1, s_2): s_1 =s_d^1}$ when learning a goal-conditioned value function $\tilde{V}_g(s;g(s_d^1))$. That is when charged with reaching the goal $g(s_d^1)$, the agent receives the reward $r(s, a) =1$ if $s' \in g(s_d^1)$ and the reward $r(s, a) =0$ otherwise. In this case, the demonstration $\tau^j = (s_0^{1,j}, \dots, s_t^{1,j})$ above consists of only the goal-relevant states, and we construct the potential via:
\begin{equation}
V_d(s_t^{1,j}) = \gamma^{H^j -t}, ~~~
\Phi^j(s) = \max_{s_t^{1,j} \in \tau^j}\left[V_d(s_t^{1,j}) + \tilde{V}_g(s; g(s_t^{1,j})) \right].
\end{equation}
The overall potential function can again be defined by $\Phi(s) = \max_j \Phi^j(s)$ to generate dense rewards with PBRS as before. In our experiments, this reward structure drives the agent to push the object to positions demonstrated by the expert. While the demonstration may not contain a sequence of states for the arm that is needed to realize the desired motion for the object, this information is baked into $\tilde{V}_g$, which (approximately) encodes sequences of states for both the object and the arm that are needed to move the object to nearby goal positions. This extension also applies to situations where a subset of the states can only be inferred through high-dimensional observations $o_t$. Indeed, the goal sets can be defined in the observation space in this case.

%% file: expirements.tex
In this section, we investigate $1)$ the speed our method learns tasks from few demonstrations and $2)$ its adaptability to sub-optimal demonstrations and high-dimensional observations. We conduct empirical studies in the following tasks: pushing a puck to a U-shaped slot, utilizing both $1)$ perfect state observations and  $2)$ imperfect, high-dimensional point cloud observations to verify the effectiveness of our proposed reward-shaping approach with binary rewards. We implement our approach on top of the off-policy Soft Actor-Critic (SAC) algorithm \citep{haarnoja2018soft}. 
See Appendix \ref{sec:app_exp} for experimental details.

\textbf{Baselines:}
We combine our reward-shaping approach with an importance sampling buffer (Sec. \ref{sec:IS}) and the annealing discount factor technique (Sec. \ref{sec:discounting}). 
The importance sampling buffer has a probability of $10\%$ to include expert demonstrations in each batch, and the annealing trick linearly increases the discount factor from $0$ to $\gamma$ during training. The label \textit{potential} refers to an agent trained with our reshaped rewards. The label \textit{buffer} refers to an agent having direct access to the expert states and actions and has included them in the importance sampling buffer as \textit{buffer}, and \textit{annealing} indicates the annealing trick in the following plots and tables. For example, Buffer + Potential + Annealing refers to an agent trained with all three modes active. 

\begin{wrapfigure}{r}{0.58\textwidth}
\subfigure[Push slot]{\includegraphics[width=0.24\textwidth]{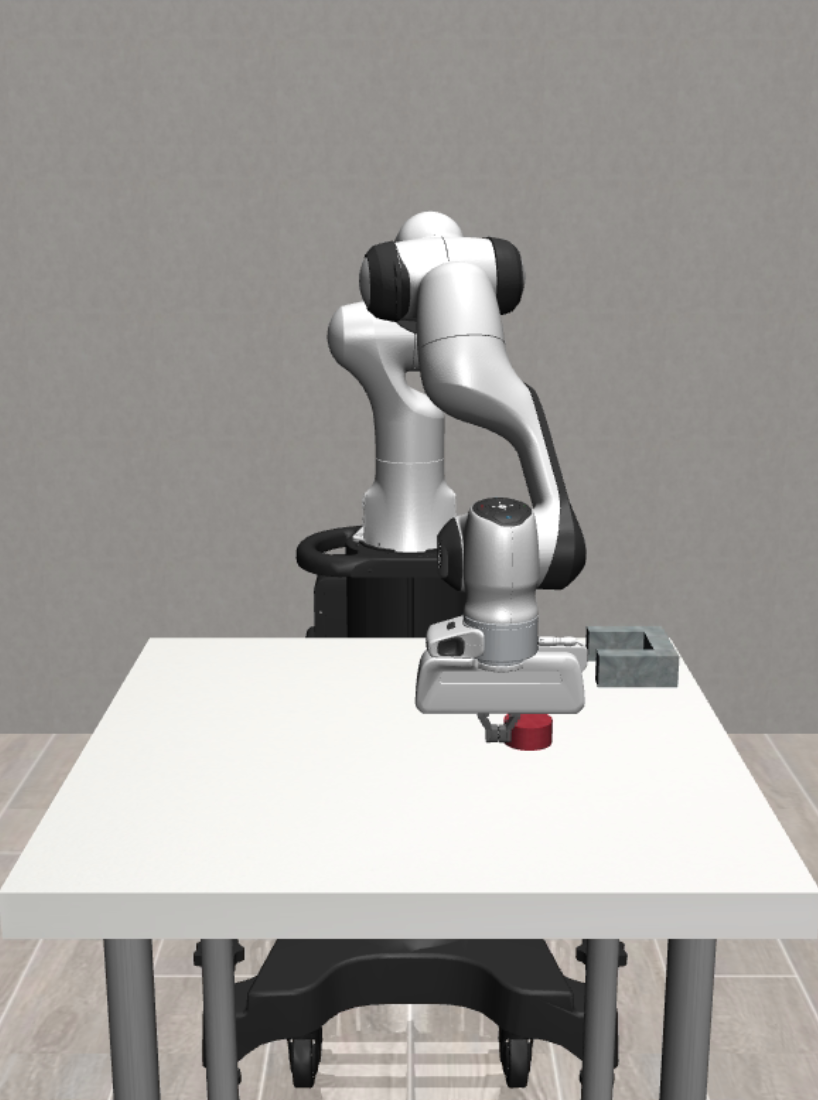}\label{fig:render_push}}
\subfigure[Potential heatmap]{\includegraphics[width=0.34\textwidth]{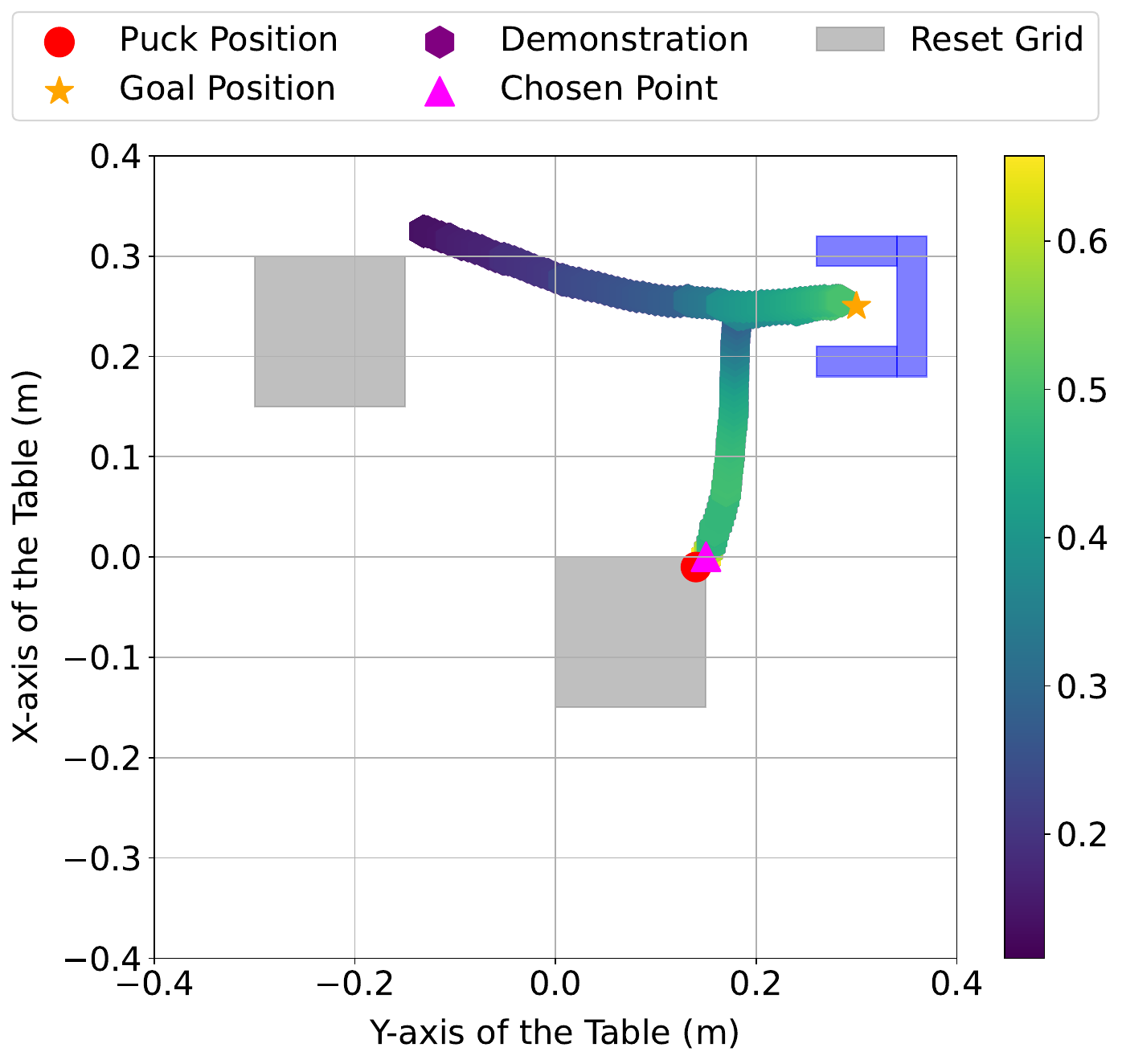}\label{fig:heatmap_push_grids}}
\vspace{-1.5em}
\caption{\small{(a) Push slot environment in Robosuite. (b) Potential heatmap for the push slot with two reset grids.}}
\vspace{-1.5em}
\label{fig:render_heatmap_push}
\end{wrapfigure}
We compared our methods with $6$ baselines: the Buffer Only mode described above, Behavior Cloning (BC) \citep{BehavioralCloning}, GAIL \citep{ho2016generative}, PWIL \citep{pwil}, DRIL \citep{dril}, and AdRIL \citep{adril}.
The Buffer Only mode does not incorporate reward-shaping and helps us to ablate how our method can improve base methods, which incorporate demonstrations into the replay buffer \citep{nair2018overcoming, rajeswaran2017learning}. 
The other baselines are standard imitation learning methods that learn from demonstrations.
We exclude comparisons of agents that solve the original sparse reward problems without reward-shaping or demonstrations, as their performance is significantly worse than all other methods tested. We implemented the baselines from \citep{arulkumaran2023pragmatic}.

\textbf{Setup:}
We created a puck-pushing environment in the Robosuite \citep{robosuite2020} simulator, where a Franka Panda robot arm aims to push a puck into a U-shaped slot that is blocked by three walls, shown in Fig. \ref{fig:render_heatmap_push}.
The puck is initialized on one or multiple reset grids (gray), and an RL agent controls the $2$-dimensional displacement of the end effector. Success occurs when the puck enters the slot. The agent observes the puck's displacement and speed relative to the end effector and the goal.
Demonstrations are collected with a sub-optimal, hand-coded expert policy that pushes the puck to predetermined anchor points around the slot to complete the task. However, as mentioned in Sec. \ref{sec:IS}, for this task, the goal sets depend only on the position of the puck but not the robot arm. Thus, we simulated situations where a human demonstrator moves the puck with their hand, and only the position of the puck is recorded in the demonstrations. 
We tested $3$ variants of the task: $1)$ the agent with perfect state estimates for the puck, with either $a)$ one reset grid or $b)$ two reset grids, and $2)$ the agent infers the puck's position from high-dimensional point-cloud observations.
The two-reset-grid variant is depicted in Fig. \ref{fig:render_heatmap_push}, and the one-reset-grid in Fig. \ref{fig:heatmap_push} in the appendix. 
The pointcloud environment has only one reset grid.
We provided \textbf{one demonstration from each active reset grid} for all experiments, as shown in Fig. \ref{fig:render_heatmap_push}.

\input{figure_latex/training_curve}
\textbf{Results:}
Fig. \ref{fig:traning_curve} shows the progression of the success rates. 
We showed the detailed statistics in the appendix (Tab. \ref{tab:mean_std}).
Learning is accelerated substantially for the agents which have access to our shaped rewards.
In contrast, previous online learning methods, GAIL, PWIL, AdRIL, and DRIL, cannot master the task.
This is because the learned value estimates $\tilde{V}_g$ encode these sorts of short-horizon pushing primitives \emph{a priori}, enabling the agent to quickly adapt to the new long-horizon task. 
Note that we do not include BC in Fig. \ref{fig:traning_curve} since it is not an online learning method, and thus no learning progress can be visualized. However, Tab. \ref{tab:mean_std} shows that while BC converges, it fails to master any pushing task even with $100$ demonstrations.
Our approaches effectively learn the tasks even with high-dimensional pointcloud observations, while other baselines fail.
Also, when successful, GAIL learns sub-optimal behaviors, requiring around $90$ steps to complete the tasks. In contrast, our approaches complete the tasks in approximately $30$ steps, showcasing their ability to learn optimal behaviors from sub-optimal demonstrations with $100$ steps.


%% file: figure_latex/training_curve.tex
\begin{figure*}[t!]
    \centering
    \includegraphics[width=0.9\textwidth]{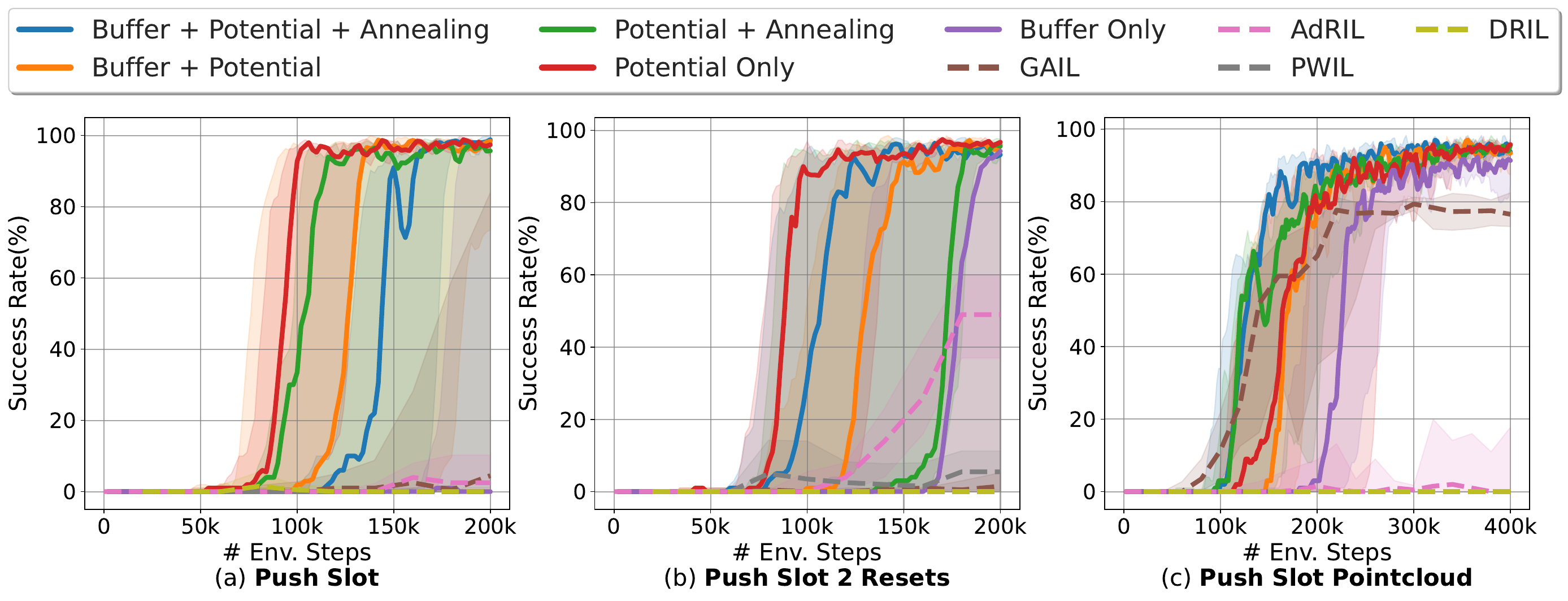}
    \vspace{-0.5em}
    \caption{\small Training progression in push slot tasks. We present the median (bold) and quartiles (shaded area) of success rates. Our approach leads to faster learning and more stable performance across tasks.}
    \vspace{-1em}
    \label{fig:traning_curve}
\end{figure*}

%% file: discussion.tex
This paper presents a general reward-shaping approach that ties together prior task-agnostic data and a small number of expert demonstrations to synthesize informative rewards for new tasks. We demonstrated that this core technique can substantially accelerate learning for new downstream tasks. However, the scope of the current study is limited in a few key ways, which leaves many avenues for future work. First, while we have demonstrated our approach with high-dimensional pointcloud observations, it will be important to demonstrate that it can extend to other rich sensory modalities, such as RGB images. Fortunately, there has been extensive work on learning and effectively leveraging latent representations and goal-conditioned value functions in those representations \citep{ma2022vip, eysenbach2019search}. Next, we are not fully leveraging the prior data set in the present framework. Several works have investigated using goal-conditioned controllers to accelerate exploration \citep{nasiriany2019planning} and mixing in samples from prior data sets when training on new tasks \citep{kumar2022pre}, rather than training from scratch on each new task. Fortunately, our reward-shaping framework is agnostic to these techniques, and the pathway to integrating our approach with these contributions leaves a clear avenue for future work.

%% file: appendix.tex
\section{Supplementary Results}
We provide the supplementary details in this section.
We learned the value estimator $\tilde{V}_g$ by training a pushing controller that can displace the puck up to $0.1$ meter in each direction.
Tab. \ref{tab:mean_std} shows the mean, standard deviation, median, and quantiles of the success rates of all methods. Fig. \ref{fig:heatmap_push} shows the heat map of pushing a puck to a slot task with one reset grid. The heatmap displays data from both state and pointcloud observation settings.

\input{table_latex/mean_success_rate}

\input{figure_latex/heatmap_push}

\newpage
\section{Experimental Details}
\label{sec:app_exp}
In this section, we present the details of the experiment. In each task, we $1)$ train the goal-conditioned value estimator $\tilde{V}_g(s;s_d)$ to move the point mass or puck to a random nearby position; $2)$ generate optimal demonstrations in maze tasks with an expert agent, and sub-optimal ones with a hand-coded policy in pushing tasks; $3)$ perform reward-shaping by substituting the original goal with an intermediate state of the demonstration with the highest potential value, as per the goal sets function $g(s_d^1)$ defined in \ref{sec:IS}; 4) train the augmented agent with the proposed reshaped reward as per \eqref{eq:reshaped_reward}. We list all the detailed settings in Tab. \ref{tab:env_para}. 

For maze tasks, the observations are the $2$-D velocity of the point $v_\text{P}$, the $2$-D velocity of the goal $v_\text{G}$, and the $2$-D displacement from the goal to the point $s_{\text{P}-\text{G}}$. The action denotes the acceleration of the point $a_\text{P}$. For pushing tasks with state observations, the observations consist of the displacement and relative velocity from the end effector to the puck $s_{\text{P}-\text{E}}, v_{\text{P}-\text{E}}$ and those from the goal to the puck $s_{\text{P}-\text{G}},  v_{\text{P}-\text{G}}$. For the pointcloud setting, as the velocities for the sampled points are not available, the observation consists only of the displacement from the end effector and the goal to the 20 sampled points $[s_{\text{P}-\text{E}}, v_{\text{P}-\text{E}}, s_{\text{P}-\text{G}},  v_{\text{P}-\text{G}}]$. The action $\Delta s_{\text{E}}$ denotes the horizontal displacement of the end effector. For all proposed approaches, we use the SAC algorithm \cite{haarnoja2018soft} from Stable-Baselines3 \cite{stable-baselines3} with the default configurations.

\input{table_latex/experiment_config}



\subsection{Pushing a Puck to a Slot}
Unless specified, the following settings include both the vector observation and pointcloud observation environments. We train the goal-conditioned value estimator $\tilde{V}_g(s;s_d)$ in an unobstructed plane. In each episode, we randomly initialize a goal in a square area of $|x,y|\leq 0.1\si{m}$ and randomly reset the end effector $0.05\si{m}$ to $0.13\si{m}$ from the puck. The goal velocity is within $|v_{\text{G},x}|,|v_{\text{G},y}|\leq0.05\si{m/s}$. The agent is rewarded $+1$ for reaching the goal and $0$ elsewhere. During training, the value estimator's outputs are clipped within the range of $[-100, 1]$, and the agent's critic values are clipped within $[-2, 1]$. In training the augmented agent, the puck is uniformly generated in the reset grid(s), and the end effector is reset randomly $0.1\si{m}$ to $0.15\si{m}$ from the puck.

\newpage
\section{Proof of Proposition \ref{prop}}
\label{sec:app_proof}
\begin{proof}
First, we recall that in the given setting the greedy agent will act at each time step by choosing a $1$-step greedy action: $\bar{\pi}^*(s) \in \arg\max_{a} r(s,a) + \gamma \Phi(f(s,a)) -\Phi(s)$. By the construction of $\Phi$ and the assumptions we impose for the result, we must have that $\Phi(s) = 1$ if $s \in G$ and $\Phi(s) <1$ otherwise, so that the potential function obtains a global maximum over the set $G$. To demonstrate that the greedy agent will reach the goal set by optimizing the the $1$-step reward, we will argue that $\Phi(s_t)$ is a strictly increasing sequence which must eventually reach the maximum value possible value of $1$. We first consider two cases: 

\textbf{Case 1:} For the first case we consider the set of states $G(1) = \set{s \in \mathcal{S} \colon \exists a \in \mathcal{A} \ \text{s.t.} \ f(s,a) \in G} \cup G$, i.e., the set of states that lie in $G$ or can reach $G$ after a single transition. On this set, because both terms $r(s,a)$ and $\gamma \Phi(f(s,a))$ in the 1-step reward are maximized by choosing actions which result in the next state reaching the goal, it much be the case that the greedy policy reaches $G$ (and the global maximum for $\Phi$) after a single transition. 

\textbf{Case 2:} For the next case, we consider all states that are more than one step away from reaching the goal, and demonstrate that we can lower-bound how much $\Phi$ increases between transitions taken by the greedy agent. To study this case, suppose we are at state $s$, will choose action $a$ and will arrive at the next state $s'$ at the next times step. We will show that we can always choose $a$ so as to increase the value of the potential during this transition by some minimum amount. We will let $s_t^j$ denote the optimal demonstration state that is chosen when constructing $\Phi(s)$ in \eqref{eq:sinle_pot}. We break down this scenario into two following subcases.

In the first subcase, suppose that we have $s = s_t^j$, so that the current state is actually at one of the demonstrated states. By construction and our assumptions, in this case we have $\Phi(s) = V_d^j(s_t^j) + \tilde{V}_g(s;s_t^j) =  V_d^j(s_t^j) + \tilde{V}_g(s_t^j;s_t^j) = V_d^j(s_t^j)+ 1$. In this case, we can choose the action taken by the demonstration at this state by setting $a = a_t^j$ so that the next state is $s' = s_{t+1}^j$. At this next state, we can lower-bound the value of the potential with $\Phi(s') = \Phi(s_d^t) > V_d^j(s_{t+1}^j) + \tilde{V}_g(s_{t+1}^j; s_{t+1}^j) = V_d^j(s_{t+1}^j) + 1$. Thus, in this case we have chosen $a$ such that $\Phi(s')$ is greater that $\Phi(s)$ by at least the difference $V_d^j(s_{t+1}^j) - V_d^{j}(s_t^j) > 0$.

In the second subcase, we suppose that $s$ is not at a state in any of the expert demonstrations. In this case we choose the action $a \in \argmax_{\tilde{a}}\tilde{V}_g(f(s,\tilde{a});s_t^j)$. That is, we choose the action which greedily follows $\tilde{V}_g$ towards the currently selected demonstration state $s_t^j$. By condition 2. in the statement of the Proposition, we then have that $\tilde{V}_g(s';s_t^j) > (1+\epsilon)\tilde{V}_g(s;s_t^j)$, which implies that we have $\tilde{V}_g(s';s_t^j) - \tilde{V}_g(s;s_t^j) > \epsilon \beta$, owing to our definition of the set of in-distribution goals in \eqref{eq:goals}. Note that $\Phi(s') \geq V_d^j(s_t^j) + \tilde{V}_g(s';s_t^j)$ and recall that we have assumed that $\Phi(s) = V_d^j(s_t^j) + \tilde{V}_g(s;s_t^j)$. Combining the facts establishes that: $\Phi(s') - \Phi(s) \geq \epsilon \beta$. 

\textbf{Completing the Proof:} We now combine the previous cases to complete the proof. Note that $\Phi(s)\geq 0$ for each $s\in \mathcal{S}$. In Case 2 above, we established that there exists $C>0$ such that for each $s \not \in G(1)$, there exists an action $a$ such that the next state satisfies $\Phi(s') - \Phi(s) >0$. This establishes that the potential $\Phi$ will monotonically increase along every trajectory generated by the agent, and that the agent will enter $G(1)$ in a finite number of steps. Then, as we established in Case 1, the agent will take at most one more step to reach $G$. This establishes the desired result.
\end{proof}

%% file: table_latex/mean_success_rate.tex
\begin{table*}[ht]
  \centering
  \scriptsize
    \begin{tabular}{cccccccc}
    \toprule
    \multirow{2}{*}{Method} & Success Rate & \multicolumn{3}{c}{Push Slot} \\
\cmidrule(lr){3-5}\cmidrule(lr){6-8} & ($\%$) & 1 Grid & 2 Grids & 1 Grid (Pointcloud) \\
    \midrule
    \midrule
    \multirow{2}{*}{Buffer + Potential + Annealing} & Mean / Std. & \textbf{99} / 0.7 & 76 / 40.2 & \textbf{95} / 3.8 \\
          & Q3 / Med. / Q1 & 100 / \textbf{100} / 99 & 98 / \textbf{96} / 85 & 97 / \textbf{95} / 94 \\
    \midrule
    \multirow{2}{*}{Buffer + Potential} & Mean / Std. & \textbf{99} / 0.8 & 77 / 40.5 & 94 / 3.4 \\
          & Q3 / Med. / Q1 & 100 / 99 / 99 & 97 / 95 / 94 & 96 / \textbf{95} / 94 \\
    \midrule
    \multirow{2}{*}{Potential + Annealing} & Mean / Std. & 69 / 47.5 & 76 / 40 & \textbf{95} / 2.2 \\
          & Q3 / Med. / Q1 & 99 / 98 / 24 & 97 / 94 / 88 & 97 / 94 / 93 \\
    \midrule
    \multirow{2}{*}{Potential Only} & Mean / Std. & 88 / 31.1 & \textbf{95} / 2.2 & 93 / 7.4 \\
          & Q3 / Med. / Q1 & 99 / 98 / 97 & 96 / 95 / 93 & 97 / \textbf{95} / 92 \\
    \midrule
    \multirow{2}{*}{Buffer Only} & Mean / Std. & 30 / 47.7 & 47 / 49.8 & 85 / 30 \\
          & Q3 / Med. / Q1 & 75 / 0 / 0 & 93 / 46 / 0 & 95 / 94 / 93 \\
    \midrule
    \multirow{2}{*}{GAIL} & Mean / Std. & 40 / 46.0 & 24 / 32.9 & 71 / 5.5 \\
          & Q3 / Med. / Q1 & 90 / 17 / 0 & 22 / 13 / 2 & 75 / 72 / 67 \\
    \midrule
    \multirow{2}{*}{PWIL} & Mean / Std. & 0.2 / 0.6 & 8.4 / 11.2 & 0.8 / 2.5\\
          & Q3 / Med. / Q1  & 0 / 0 / 0 & 11.3 / 5.5 / 0.5 & 0 / 0 / 0\\
    \midrule
    \multirow{2}{*}{DRIL} & Mean / Std. & 1 / 2.2 & 0 / 0 & 0.1 / 0.3\\
          & Q3 / Med. / Q1 & 0.75 / 0 / 0 & 0 / 0 / 0 & 0 / 0 / 0\\
    \midrule
    \multirow{2}{*}{AdRIL} & Mean / Std. & 6.1 / 8.0 & 46.5 / 19.1 & 20.8 / 36.4\\
          & Q3 / Med. / Q1 & 10.25 / 2.5 / 0 & 60.5 / 50 / 40.5 & 17.8 / 0 / 0\\
    \midrule
    \multirow{2}{*}{BC 1 Demo $^*$} & Mean / Std. & 1 / 1.3 & 1 / 0.7 & 0 / 0.3 \\
          & Q3 / Med. / Q1 & 1 / 0 / 0 & 1 / 0 / 0 & 0 / 0 / 0 \\
    \midrule
    \multirow{2}{*}{BC 10 Demos} & Mean / Std. & 36 / 12.9 & 13 / 5.3 & 0 / 0 \\
          & Q3 / Med. / Q1 & 47 / 35 / 26 & 16 / 12 / 9 & 0 / 0 / 0 \\
    \midrule
    \multirow{2}{*}{BC 100 Demos} & Mean / Std. & 51 / 8.0 & 22 / 9.6 & 0 / 0.5 \\
          & Q3 / Med. / Q1 & 54 / 51 / 48 & 26 / 23 / 17 & 1 / 0 / 0 \\
    \bottomrule
    \end{tabular}%
  \caption{\small{Mean, standard deviation, median, and quantiles of success rates. 
    The statistics are calculated over $10$ agents trained for the corresponding total environment steps (Tab. \ref{tab:env_para}).
    Our potential shaping approach results in more stable performance than the other baselines. ($^*$ For 2/3 Reset Grids, the same number of demonstrations are provided.)}}
  \vspace{-1.5em}
  \label{tab:mean_std}
\end{table*}%

%% file: figure_latex/heatmap_push.tex
\begin{figure}[ht!]
    \centering
    \includegraphics[width=0.5\textwidth]{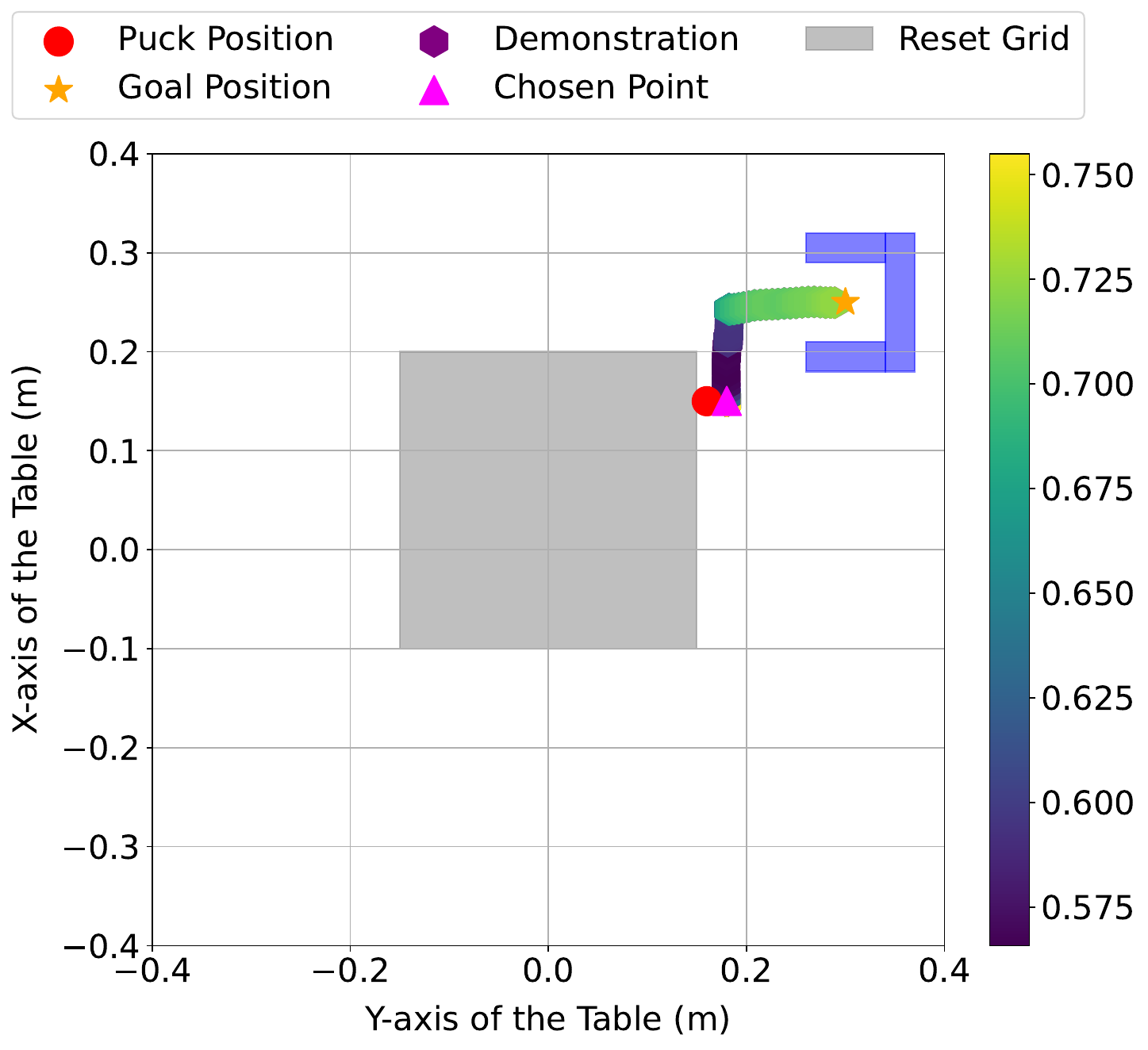}
    \caption{\small{Potential heatmap for the push slot with single reset grid.
    The agent constructs the $V_d^{j}(s_t^j) + \tilde{V}_g(s;s_j^t)$ in \eqref{eq:demo_pot} and \eqref{eq:sinle_pot} from a fixed state $s$ (green) and different demonstrated states (purple to yellow).
    The state chosen by the maximizations in \eqref{eq:sinle_pot} when constructing $\Phi$ is marked in pink.
    }}
    \label{fig:heatmap_push}
\end{figure}

%% file: table_latex/experiment_config.tex
\begin{table*}[thbp]
  \centering
  \scriptsize
    \setlength{\extrarowheight}{2pt}
    \begin{tabular}{|c||c|c|c|}
    \hline
    Environment & 1 Grid & 2 Grids & 1 Grid (Pointcloud) \\\hline
    Observation & \multicolumn{2}{c|}{$[s_{\text{P}-\text{E}}, v_{\text{P}-\text{E}}, s_{\text{P}-\text{G}},  v_{\text{P}-\text{G}}]\in\mathbb{R}^{8}$} & $[s_{\text{P}_{1:20}-\text{E}}, s_{\text{P}_{1:20}-\text{G}}]\in\mathbb{R}^{20\times 4}$ \\\hline
    Action & \multicolumn{3}{c|}{$\Delta s_{\text{E}}\in\mathbb{R}^2$} \\\hline 
    Discount Factor $\gamma$ & \multicolumn{3}{c|}{$0.99$}  \\\hline
    \# Demos $K$ & $1$ & $2$ & $1$ \\\hline
    Demo Steps $H$ & $125$ & $146, 143$ & $125$ \\\hline
    Annealing Steps $M$ & \multicolumn{3}{c|}{$10$k} \\\hline
    \# Total Env. Steps & \multicolumn{2}{c|}{$200$k} & $400$k \\\hline
    Success Distance Threshold ($L_2$ norm) & \multicolumn{3}{c|}{$0.02\si{m}$}\\\hline
    Success Velocity Threshold ($L_2$ norm) & \multicolumn{2}{c|}{$0.02\si{m/s}$} & {$0.1\si{m/s}$} \\\hline
    Batch Size & \multicolumn{3}{c|}{$512$} \\\hline
    Replay Buffer Size & \multicolumn{3}{c|}{$1$M} \\\hline
    Gradient Steps of RL Agent & \multicolumn{3}{c|}{$2$} \\\hline
    \end{tabular}%
  \caption{\small{Comprehensive task configurations.
  }}
  \label{tab:env_para}%
\end{table*}%